\documentclass{article}

\usepackage[nonatbib,final]{neurips_2024}

\usepackage[utf8]{inputenc} % allow utf-8 input
\usepackage[T1]{fontenc}    % use 8-bit T1 fonts
\usepackage{url}            % simple URL typesetting
\usepackage{booktabs}       % professional-quality tables
\usepackage{amsfonts}       % blackboard math symbols
\usepackage{nicefrac}       % compact symbols for 1/2, etc.
\usepackage{microtype}      % microtypography

\usepackage{graphicx}
\usepackage{algorithm}
\usepackage{algpseudocode}
\usepackage{amsmath}
\usepackage{subcaption}
\usepackage{mathtools}
\usepackage{hyperref}       % hyperlinks
\usepackage{xcolor}         % colors
\usepackage[numbers]{natbib}

\title{AdaEDL: Early Draft Stopping for Speculative Decoding of Large Language Models via an Entropy-based Lower Bound on Token Acceptance Probability}

\makeatletter
\let\@fnsymbol\@arabic
\makeatother

\author{%
  Sudhanshu Agrawal \qquad Wonseok Jeon \qquad Mingu Lee \\
  Qualcomm AI Research\thanks{Qualcomm AI Research is an initiative of Qualcomm Technologies, Inc.} \\
  \texttt{\{sudhagra, wjeon, mingul\}@qti.qualcomm.com} \\
}

\begin{document}

\maketitle

\begin{abstract}
  Speculative decoding  \cite{spd-leviathan} is a powerful technique that attempts to circumvent the autoregressive constraint of modern Large Language Models (LLMs). The aim of speculative decoding techniques is to improve the average inference time of a large, \textit{target} model without sacrificing its accuracy, by using a more efficient \textit{draft} model to propose draft tokens which are then verified in parallel. The number of draft tokens produced in each drafting round is referred to as the draft length and is often a static hyperparameter chosen based on the acceptance rate statistics of the draft tokens. However, setting a static draft length can negatively impact performance, especially in scenarios where drafting is expensive and there is a high variance in the number of tokens accepted. \textbf{Ada}ptive \textbf{E}ntropy-based \textbf{D}raft \textbf{L}ength (\textbf{AdaEDL}) is a simple, training and parameter-free criteria which allows for early stopping of the token drafting process by approximating a lower bound on the expected acceptance probability of the drafted token based on the currently observed entropy of the drafted logits. We show that AdaEDL consistently outperforms static draft-length speculative decoding by 10\%-57\%  as well as other training-free draft-stopping techniques by upto 10\% in a variety of settings and datasets. At the same time, we show that AdaEDL is more robust than these techniques and preserves performance in high-sampling-temperature scenarios. Since it is training-free, in contrast to techniques that rely on the training of dataset-specific draft-stopping predictors, AdaEDL can seamlessly be integrated into a variety of pre-existing LLM systems. 
\end{abstract}

\section{Introduction}
   Large Language Models (LLMs) have been shown to have impressive performance on a variety of tasks including creative writing, summarization, and translation \cite{brown2020languagemodelsfewshotlearners}. In particular, in recent years, several \textit{foundation models} such as Llama2 \cite{touvron2023llama2openfoundation}, Llama3 \cite{dubey2024llama3herdmodels},  GPT-4 \cite{openai2024gpt4technicalreport}, and Claude \cite{bai2022constitutionalaiharmlessnessai} have shown to exceed expectations and generalize to coding \cite{rozière2024codellamaopenfoundation}, display agentic abilities \cite{liu2023agentbenchevaluatingllmsagents}, interpret images \cite{liu2023visualinstructiontuning}, and more. In all such systems, an LLM's job remains the same - prediction of the next token via autoregressive generation. Autoregressive generation is fundamentally sequential in nature, since the prediction of the next token can only occur once the previous token has been generated. This reduces the ability of an LLM to parallelize, creating a bottleneck in the maximum number of generated tokens per second (TPS). 

    Speculative decoding techniques attempt to introduce parallelism to this system. Consider a scenario where the objective is to perform inference on a \textit{target model}, say Llama2-7B. A smaller \textit{draft model} with the same tokenizer as the target model, say TinyLLama-1B \cite{zhang2024tinyllamaopensourcesmalllanguage} is then chosen. Given a prompt, the draft model is allowed to run autoregressively, producing a set of candidate draft tokens. These tokens are then consumed by the target model which produces logits for each draft token, representing a probability distribution. Rejection sampling \cite{spd-leviathan} then guarantees that the draft tokens which are \textit{accepted} via this process will preserve the distribution of the original target model. Thus, by running a small model autoregressively and a large model in parallel, the system as a whole experiences an increase in average token rate. 

\begin{figure}[h]
        \centering
        \includegraphics[width=1\linewidth]{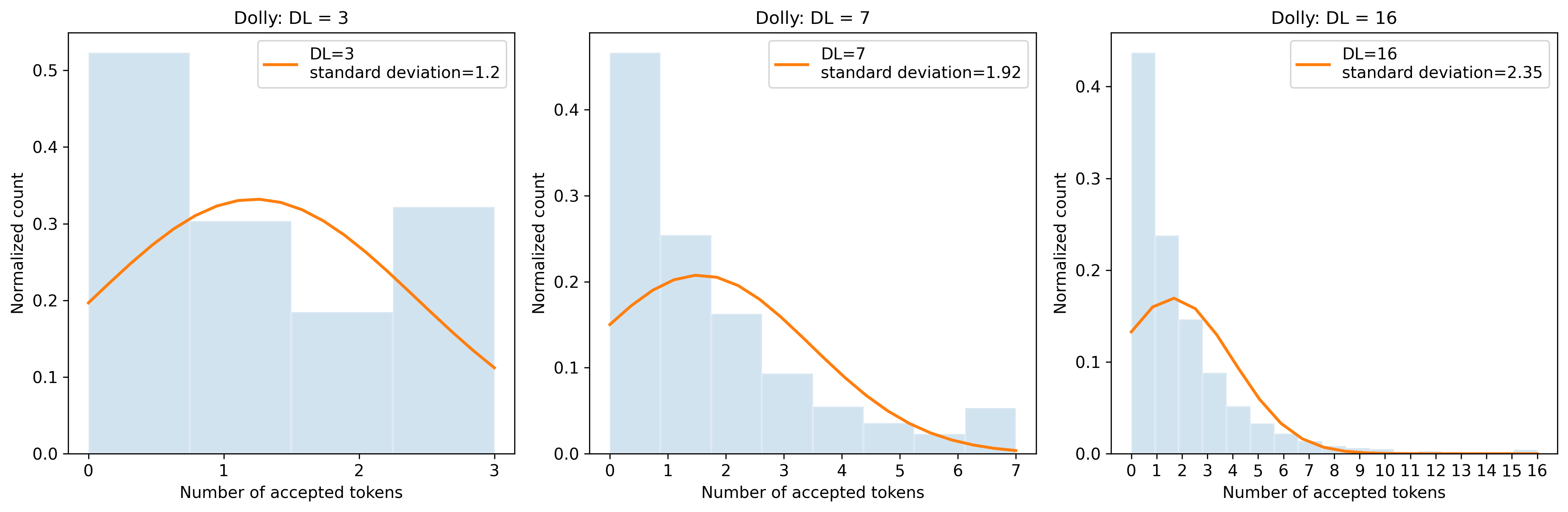}
        \caption{The number of accepted tokens across drafting rounds displays a high variance, leading to under or over-utilization of the draft model in static draft length speculative decoding methods.}
        \label{fig:dolly-acceptance}
    \end{figure}

    However, a limiting factor of this system is that the \textit{number} of draft tokens produced, referred to as the draft length (DL), if fixed over multiple rounds of drafting, can reduce the average token rate. This may be caused due to over-utilization of a poorly performing draft model, or symmetrically, because the draft model is under-utilized which does not allow the speculative decoding system to reach its maximum possible performance. For example, since most target models are large foundation models, designed to have high accuracy on a variety of tasks, it is likely that a smaller draft model finetuned to match the target model distribution for a particular task may have varying levels of accuracy when the target model switches tasks. 
    In such scenarios, one can see a high variance in the number of accepted draft tokens. That is, in some drafting rounds, almost all the draft tokens are accepted, whereas in some, almost all are rejected. Figure \ref{fig:dolly-acceptance} considers the creative writing task from the Dolly-15k dataset \cite{DatabricksBlog2023DollyV2}. For this dataset, we see that for a standard speculative decoding system operating with various static draft lengths, the number of accepted tokens follows a normal distribution, with $\text{num-accepted-tokens}$ taking on almost every value from $0 \text{ to} \text{ max-draft-length}$ with varying frequency. We include similar figures for the CNN-DM (summarization) \cite{see-etal-2017-get} and WMT-19 (German-English translation) \cite{wmt19translate} datasets along with the details to set up these speculative decoding systems in Appendix \ref{appendix-acceptance-rates} Figures \ref{fig:cnn-acceptance}, \ref{fig:wmt-acceptance}.

    This motivates the need for an \textbf{adaptive draft length} speculative decoding system where the draft length at every drafting round can be determined on-the-fly. AdaEDL approaches this problem with a \textit{go-no-go strategy}: while drafting tokens, at every iteration, AdaEDL establishes a draft-stopping criteria by approximating a lower bound on the expected acceptance probability of the drafted token by using the entropy of the draft model logits at that iteration. If the criteria is satisfied, drafting stops and verification by the target model is performed. We provide a theoretical basis to our proposed draft-stopping criteria formulation, deriving how it relates to the acceptance probability of the draft model. To validate our approach, we perform experiments across various maximum draft length settings, across multiple datasets and sampling temperature settings, and for various target and draft model choices, showing that this new draft-stopping strategy is more effective and robust than draft-stopping strategies which simply use the probability value of the most-likely token as a draft-stopping criteria - for example, those used in BiLD \cite{kim2023speculativedecodingbiglittle} and Draft \& Verify \cite{zhang2024draftverifylossless}. Indeed, AdaEDL may be used to complement either of these algorithms. At the same time, our proposed system avoids the need to train an independent network to act as a binary classifier for early draft-stopping for a specific model and task, such as the approaches followed by SpecDec++  \cite{huang2024specdecboostingspeculativedecoding} and DISCO \cite{mamou2024dynamicspeculationlookaheadaccelerates}, which makes AdaEDL a simple and straightforward improvement to boost the token rate of speculative decoding LLM systems.

    % Finally, we show that when the decision of whether to draft a particular token or not is viewed as a binary classification problem, AdaEDL shows a high correlation to the F1 score of this classification problem. 
  
  % 1. Introduction to the problem 
  % 2. Introduction to speculative decoding and the high variance observed in static draft lengths and high temperatures 
  % 3. Talk about the acceptance rate as a target metric that you want to optimize for. 
  % 4. Talk about how you by creating an approximate lower bound on the acceptance rate, you can derive a criteria using just entropy of the draft model and use that as a draft stopping criteria 
  % 5. Mention the gains in 1) expensive draft models, 2) baseline performance 3) high temperature performance 4) controllability and predictability 

\section{Problem setting and existing methods} \label{sec:existing-methods}
    Following a notation similar to the original speculative decoding formulation, let $TM$ be the target model whose inference we are trying to accelerate. Let $DM$ be a more efficient approximation of this target model, referred to as the draft model. Let us denote the $t^{th}$ token in the prompt by $x_t$. Then the probability distribution of $TM$ at any given token $x_t$ may be denoted as $p_{TM}(x_t | x_{<t})$. Similarly, the probability distribution of the draft model, $DM$ may be denoted as $p_{DM}(x_t | x_{<t})$. As noted in \cite{spd-leviathan}, several strategies such as nucleus sampling, top-k sampling, and others may each be viewed as sampling from an adjusted probability distribution. For notational simplicity, we refer to these probabilities, adjusted or not, as $p_{TM} (x) $ and $p_{DM} (x)$. In this notation, we can now describe the various decoding techniques: 

    \paragraph{Autoregressive decoding} In this baseline technique, we only consider the target model and at every iteration, a new token is sampled as $$x_t \sim p_{TM}( \cdot | x_{<t}) $$

    \paragraph{Speculative decoding} The system first allows a draft model to consume the tokens $x_{<t}$. The draft model then autoregressively generates $L$ draft tokens $d_1, d_2, \cdots d_L$, where $L$ is the maximum draft length. This can be viewed as sampling a token $$ d_i \sim p_{DM}( \cdot | x_{<t}, d_1, \cdots, d_{i-1}) $$ After drafting, the target model evaluates these draft tokens in parallel, producing probabilities $p_{TM}(x)$. The draft token $d_i$ is \textit{accepted} if $p_{DM}(x) \leq p_{TM}(x)$, producing a token $x_i = d_i$. Otherwise, it is \textit{rejected} with probability $1 - p_{TM}(x)/p_{DM}(x)$ and a new token $x_i$ is sampled from an adjusted distribution $x_i \sim p_{TM}'(x) = norm(max(0, p_{TM}(x) - p_{DM}(x))) $. A token $x_i$ sampled via this method, referred to as rejection sampling, is guaranteed to follow the probability distribution of the original target model, that is, $x_i \sim p_{TM}(x) $ \cite{spd-leviathan}. 

    To further improve this process, the draft length $L$ may be made \textit{adaptive}, to avoid the cost of drafting tokens that are likely to be rejected anyway. 

    \paragraph{Adaptive draft length via maximum confidence speculative decoding} Some systems consider the top-1 probability of the drafted logits. That is, at a given drafting stage, the system considers the token with the highest probability $\max_x p_{DM} (x | x_{<{t}}, d_1, \cdots, d_{i-1})$. If this value, the \textit{maximum confidence} among any possible token, is less than a threshold, $\lambda$, the system is considered to be \textit{under-confident} and drafting stops. $$\max_x p_{DM}(x) < \lambda $$
    This is a simple method to avoid wastage during the drafting phase and is effectively employed alongside other techniques in Draft \& Verify \cite{zhang2024draftverifylossless} and BiLD \cite{kim2023speculativedecodingbiglittle}.  
    However, this scheme fails to take into account the overall probability distribution during drafting. 
    %Consider a degenerate scenario where for a vocabulary of size 4, the probabilities of various tokens are all exactly equal $[0.25, 0.25, 0.25, 0.25]$. In this scenario, if the stopping threshold is set to $\lambda = 0.2$, the system would continue to draft, even though it's clear that the \textit{entropy} of the system is high.%
    This motivates the need for a new drafting stopping mechanism that considers the probabilities of all possible tokens while making a go-no-go decision. 

    \paragraph{Adaptive draft length via a trained predictor} Several works involve the training of a small network to act as a predictor to determine optimal draft lengths, for example, SpecDec++ \cite{huang2024specdecboostingspeculativedecoding} which trains a ResNet and DISCO \cite{mamou2024dynamicspeculationlookaheadaccelerates} which trains a FFN as a predictor for early draft stopping. AdaEDL is distinct from these method and specifically aims to be training and parameter-free, similar to draft stopping via maximum confidence, to allow for greater generalization across datasets and models. 

\section{Adaptive entropy-based draft length speculative decoding (AdaEDL) } \label{sec:adaedl-method} 
\begin{figure}
        \centering
        \includegraphics[width=1\linewidth]{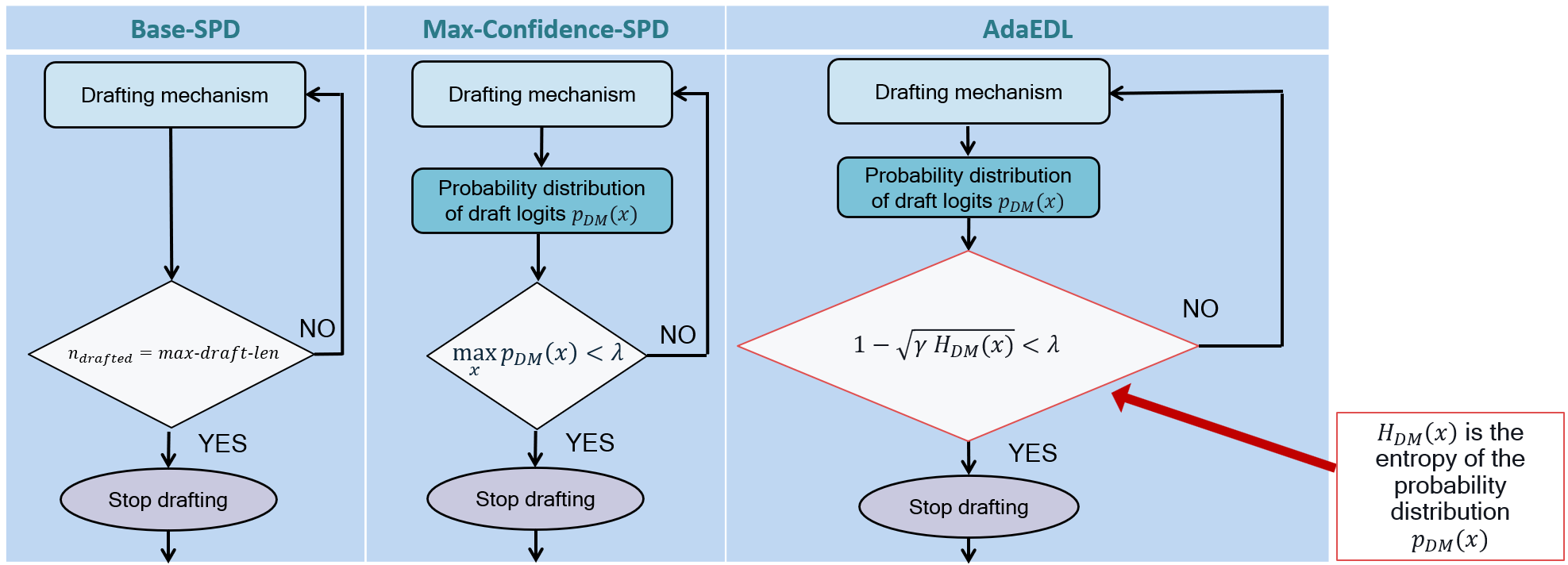}
        \caption{AdaEDL performs adaptive early-draft-stopping via an entropy-based lower bound on the expected acceptance rate. }
        \label{fig:method-flowcharts}
    \end{figure}

    AdaEDL operates in the same setup as the above adaptive draft length techniques but instead, establishes a stopping criteria using an entropy-based lower bound on the token acceptance probability. If we consider the probability distribution of the draft model, $p_{DM}(x)$ and its corresponding entropy $H_{DM}(x)$, we see that  $ 1 - \sqrt{\gamma H_{DM}(x)} $ serves as an approximate lower bound on the expected acceptance rate. This allows us to formulate a drafting system where drafting is stopped if this criteria falls below a threshold $ \lambda $,  indicating that the expected acceptance rate is also below this threshold: $$ 1 - \sqrt{\gamma H_{DM}(x)} < \lambda$$ 
    Here, $\gamma > 0$ is a hyperparameter. We include a detailed derivation of this lower-bound approximation in Appendix \ref{appendix-entropy-derivation}. Additionally, we visualize AdaEDL alongside the baseline speculative decoding systems mentioned above in Figure \ref{fig:method-flowcharts}, highlighting the novel early draft-stopping criteria we introduce. 

    \paragraph{Improving $\lambda$ via dynamic updates} \label{sec:threshold-update-algorithm} The stopping threshold $\lambda$ can be further optimized by making it responsive to the current acceptance rate statistics of the system. In particular, we aim to achieve a target acceptance rate $\alpha$ - increasing the stopping threshold if it is not currently being reached and decreasing it otherwise. By maintaining an exponential moving average of the acceptance rate and using it to update the stopping threshold in this manner, we can achieve better performance over longer generations. We modify the threshold update strategy described in \cite{zhang2024draftverifylossless} and describe it in Algorithm \ref{alg:dynamic-lambda}.

    \begin{algorithm} [t]
        \caption{Dynamic updates for stopping threshold $\lambda$ }\label{alg:dynamic-lambda}
        \begin{algorithmic}[1]
            \State $L \gets \text{max\_draft\_length}$ 
            \State $n_{drafted} \gets \text{num\_drafted\_tokens}$
            \State $n_{acc} \gets \text{num\_accepted\_tokens}$
            \State $i \gets \text{cur\_drafting\_round}$ 
            \State $\lambda \gets \text{cur\_stopping\_threshold}$
            \While{ $ i \geq 1 $ } 
                \State $AR_i \gets $ $n_{acc}/n_{drafted}$ \Comment{Calculate the current acceptance rate}
                \State $AR \gets \beta_1 AR + (1-\beta_1)AR_i$ 
                \If{$AR < \alpha$} \Comment{Not yet meeting the target acceptance rate}
                   \State $\lambda' \gets \lambda + \epsilon$  
               \ElsIf{$n_{acc} \neq L$}  \Comment{Not yet drafting max possible tokens}
                    \State $\lambda' \gets \lambda - \epsilon$  
                \Else                 
                    \State $\lambda' \gets \lambda$ 
                \EndIf
            \State $\lambda \gets \beta_2 \lambda + (1-\beta_2)\lambda'$ \Comment{Update the stopping threshold}
    \EndWhile
    \end{algorithmic}
    Hyperparameters: $\alpha$ = target acceptance rate, $\epsilon$ = step size, $\beta_1$, $\beta_2$ are used to calculate the exponential moving averages. Refer to Section \ref{sec:hyperparameters} for additional details.
    \end{algorithm}

    We further discuss the hyperparameters introduced here, $\gamma$, $\lambda$, $\beta_1$, $\beta_2$, $\alpha$, and $\epsilon$, their typical values and ranges, as well as the sensitivity of AdaEDL to these hyperparameters in Section \ref{sec:hyperparameters}.

    % \begin{itemize}
    %     \item Let $AR_i$ be the acceptance rate at the current drafting round. 
    %     \item The moving average is updated as $AR \leftarrow \beta_1 AR + (1- \beta_1)AR_i$, where $\beta_1$ is a hyperparameter. 
    %     \item We set a target acceptance rate, $\alpha$, and if the $AR \leq \alpha $, we set $\lambda' = \lambda + \epsilon$, where $\epsilon$ is a step size hyperparameter. 
    %     \item Otherwise, 
    %     \begin{itemize}
    %         \item if we are already drafting the maximum number of tokens, $\lambda' = \lambda $ 
    %         \item else, $\lambda' = \lambda - \epsilon $
    %     \end{itemize}
    %     \item Finally, $\lambda \leftarrow \beta_2 \lambda + (1-\beta_2)\lambda'$, where $\beta_2$ is a hyperparameter. 
    % \end{itemize}

    % 1. The basic variables involved in SPD, draft tokens, target model, draft model (could be finetuned \cite{goel}, a tree \cite{wonseok}, a subset of the layers of the original model \cite{draft and verify}, a cascading set of draft models \cite{},  probabilities etc. 
    % 2. Show the naive method of early stopping via max confidence (define it), and mention the papers that do this 
    % 3. State your new formulation 
    % 4. Try to derive how you got to this formulation (could be moved to appendix) 
    % 5. Flowchart diagram 

\section{Experimental Results} \label{sec:experiments}
    In all of the following results, \textbf{AdaEDL} refers to our proposed method, described in in Section \ref{sec:adaedl-method} with dynamically updated stopping thresholds as described by Algorithm \ref{alg:dynamic-lambda}. \textbf{Max-Confidence-SPD} refers to a speculative decoding system which implements the early draft-stopping strategy based on the token with the highest probability described in Section \ref{sec:existing-methods}, which is further improved by using the same dynamic threshold strategy to have a fair comparison with AdaEDL. \textbf{Base-SPD} refers to vanilla speculative decoding with no early draft-stopping strategy employed and \textbf{Autoregressive} refers to standard autoregressive decoding of LLMs which is the baseline that speculative decoding systems attempt to improve on. 
    
    We evaluate the performance of AdaEDL across various datasets and tasks: Dolly-15k (creative writing) \cite{DatabricksBlog2023DollyV2}, WMT-19 (German-English translation) \cite {wmt19translate}, and CNN-DM (summarization) \cite {see-etal-2017-get}. The entire Dolly test dataset (708 samples), the entire WMT-19 test dataset (600 samples), and the first 1000 samples of the CNN-DM test dataset were used to ensure a large sample size. All token rate (TPS) numbers reported are calculated as the averages over all samples in each dataset. All experiments were performed on a single NVIDIA A100 with 80GB of GPU memory in FP32 precision. We acknowledge here that results may vary on different hardware which may lead to faster or slower inference of a draft or target model and in particular, their relative speed. We discuss this point in Appendix \ref{par:hardware-impact}. 
    
     We show in the following experiments that AdaEDL will always either exceed or match the performance of other decoding systems while also proving less sensitive to chosen hyperparameters and robust to the system's settings. 

    \subsection{Performance with fast finetuned draft models} \label{sec:fast-finetuned-draft}
         In the following experiments, we use Llama2-7B as the target model with sampling temperature set to 0.7. The draft model, Llama2-Drafter-115M, is a 115M parameter model, distilled and finetuned using direct alignment \cite{goel2024directalignmentdraftmodel} to closely match the target model distribution . As a result, even Base-SPD results in higher acceptance rates than one would observe from an off-the-shelf draft model.
         
         In Table \ref{tab:fast-finetuned-draft}, we indicate the performance of AdaEDL across the various datasets and tasks mentioned above - Dolly-15k, WMT-19, and CNN-DM and report the average token rate on each of these datasets. We observe that AdaEDL consistently matches or beats both Max-Confidence-SPD and Base-SPD on all tasks by a competitive margin. In particular, AdaEDL has a significant advantage in the open-ended CNN-DM summarization task. 
    
        \begin{table} [htbp]
            \caption{Performance of AdaEDL vs Max-Confidence-SPD vs Base-SPD for various maximum draft lengths and datasets. Target model = Llama2-7B, draft model = Llama2-Drafter-115M, sampling temperature = 0.7.}
            \label{tab:fast-finetuned-draft}
            \centering
            \begin{tabular}{cccc}
               \hline
               Max DL = $16$  & CNN-DM  &  Dolly-15k  &  WMT-19 \\
               \hline 
               Autoregressive  & $25.74$ & $29.02$ & $29.80$  \\ 
               Base-SPD  &  $36.30$ & $32.10$  &  $22.30$ \\
               Max-Confidence-SPD  & $49.50$ & $55.80$ & $43.70$ \\
               AdaEDL  & $\mathbf{54.10}$ & $\mathbf{56.10}$  & $\mathbf{43.90}$ \\
              \hline
              Max DL = $7$  & CNN-DM  &  Dolly-15  &  WMT-19 \\
               \hline 
              Autoregressive  & $25.74$ & $29.02$ & $29.80$ \\ 
               Base-SPD  & $51.50$  & $47.60$  & $32.70$ \\
               Max-Confidence-SPD  &  $53.50$ & $56.60$  & $45.10$ \\
               AdaEDL  & $\mathbf{56.90}$ & $\mathbf{57.10}$ & $\mathbf{45.20}$ \\
               \hline
              Max DL = $3$  & CNN-DM  &  Dolly-15k  &  WMT-19 \\
               \hline 
              Autoregressive  & $25.74$ & $29.02$ & $29.80$ \\ 
               Base-SPD  & $54.10$ & $54.70$ & $40.90$ \\
               Max-Confidence-SPD  & $53.10$ & $55.70$  & $42.50$ \\
               AdaEDL  & $\mathbf{55.70}$ & $\mathbf{55.80}$ & $\mathbf{45.00}$ \\
            \hline \\
            \end{tabular}
    
        \end{table}
        
    % Show results for temperature = 0.7 for various DL's (3, 7, 16) for various datasets (Dolly-15k, WMT-19, CNN-DM)

    \subsection{Performance of AdaEDL with expensive draft models} \label{sec:expensive-draft}
    A smaller draft model may be faster at drafting, but sacrifices acceptance rate. Symmetrically, a larger draft model may lead to higher acceptance rates, but lower overall inference speed due to its own cost of inference. 
        \subsubsection{Pythia}
        The Pythia family of models \cite{biderman2023pythiasuiteanalyzinglarge} consists of models of sizes ranging from 70M to 12B parameters. As a result, they represent an ideal set of models to test the performance of various decoding systems when the ratio of the target model size to the draft model size is varied - that is, when $\frac{|TM|}{|DM|}$ is varied. 
        
\begin{figure} 
            \centering
            \includegraphics[width=1\linewidth]{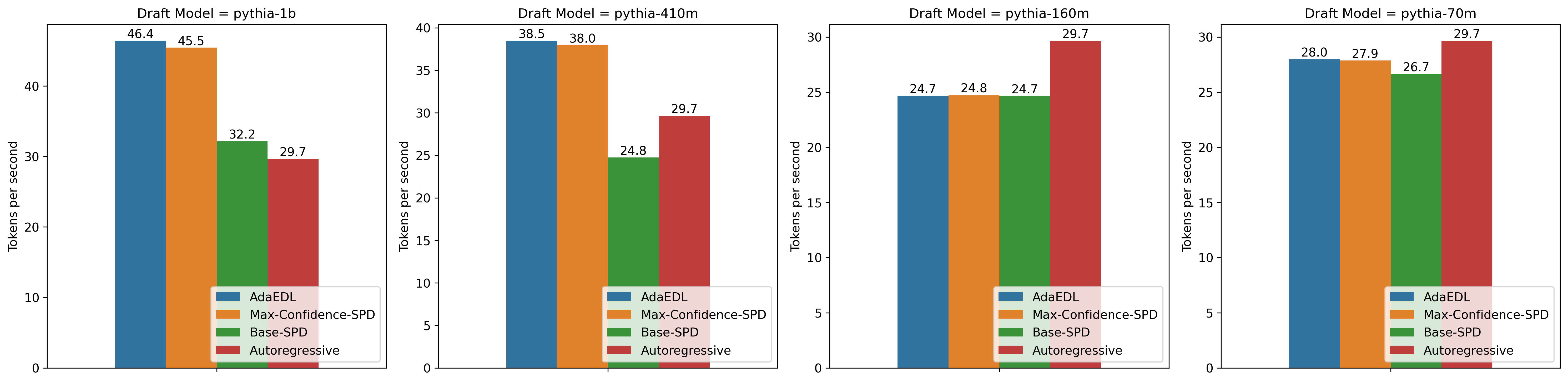}
            \caption{Target model = Pythia-6.9B with various ${|TM|}/{|DM|}$ ratios demonstrates that AdaEDL opens up the possibility of using speculative decoding with much larger draft models. Max draft length = $7$, sampling temperature = $1.0$, dataset = CNN-DM. }
            \label{fig:pythia-tm-dm}
        \end{figure}

        Figure \ref{fig:pythia-tm-dm} sets Pythia-6.9B as the target model and considers the performance of Pythia-1B, Pythia-410M, Pythia-160M, and Pythia-70M as draft models for the CNN-DM (summarization) task. We demonstrate that adaptive draft length techniques enable us to use upto $10\times$ larger draft models in a scenario where otherwise, the maximum token rate achievable with Base-SPD would have been $32.2$ tokens per second (TPS). We see that when Pythia-160M or Pythia-70M is used as the draft model, autoregressive decoding outperforms all speculative decoding methods with a token rate of $29.7$ TPS. Without adaptive draft length techniques, the maximum token rate achievable through Base-SPD is $32.2$ TPS with Pythia-1B as a draft model for an $8\%$ speedup. However, when Pythia-1B is used to draft for Pythia-6.9B with AdaEDL, the system achieves a token rate of $46.4$ TPS for a $56\%$  speedup. Thus, AdaEDL enables speculative decoding in a scenario where autoregressive decoding would normally be the candidate method. This also opens up the possibility of \textit{finetuning} larger draft models, which would presumably lead to higher acceptance rates without sacrificing performance if AdaEDL were to be enabled. We believe that this is a promising direction of investigation for future work.  

        \subsubsection{TinyLlama}
\begin{figure} 
            \centering
            \includegraphics[width=0.50\linewidth]{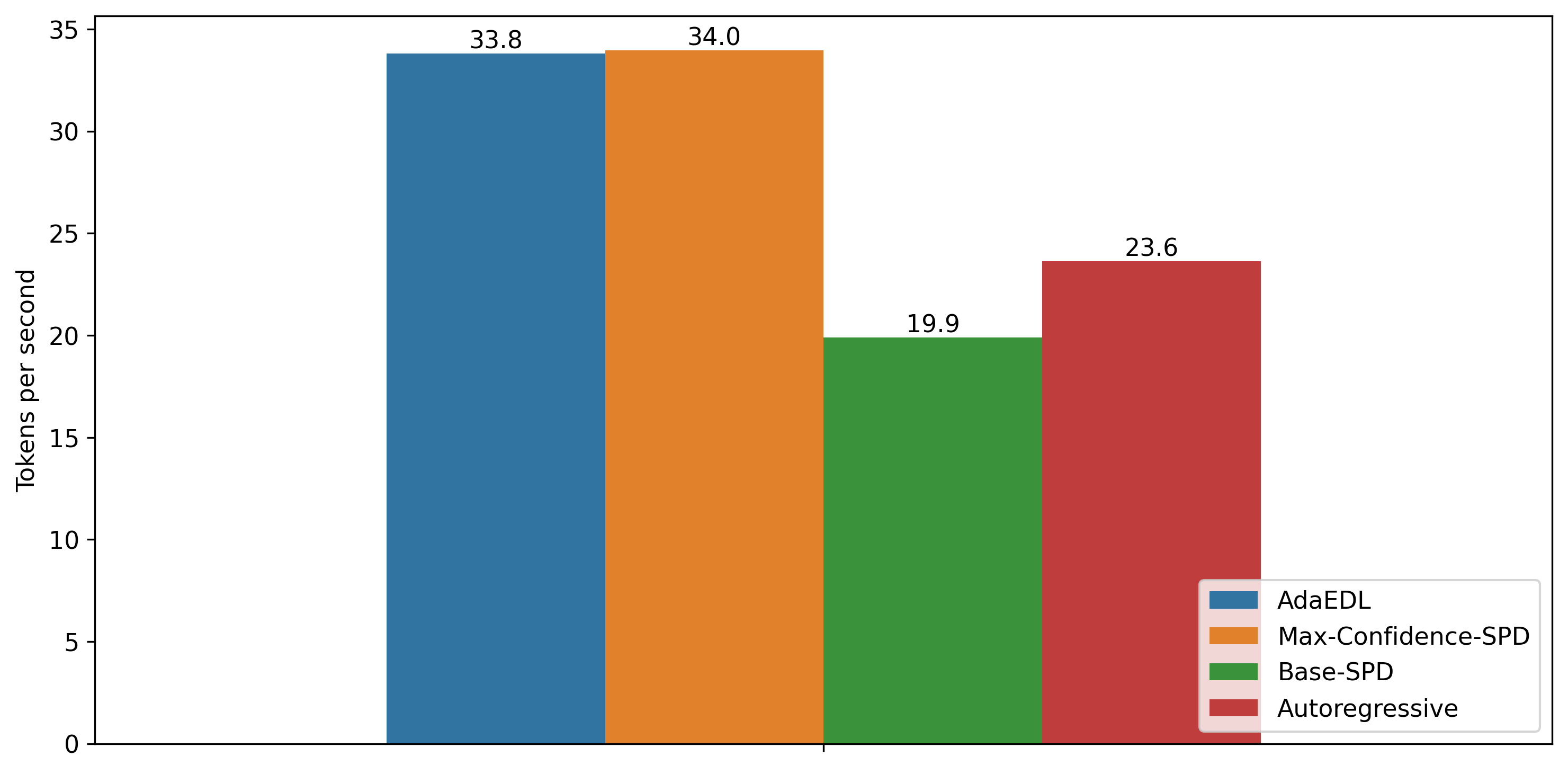}
            \caption{Target model = LLama2-7B, draft model = TinyLLama-1B, max draft length = $7$, sampling temperature = $1.0$,  when used in an adaptive draft length decoding scheme, surpass autoregressive and Base-SPD performance. Dataset = CNN-DM. }
            \label{fig:tinyllama}
        \end{figure}

        Motivated by the results in Section \ref{sec:expensive-draft}, Figure \ref{fig:tinyllama} shows the performance of various decoding methods when the target model is Llama2-7B and the draft model is a standard 1B model - TinyLlama-1B \cite{zhang2024tinyllamaopensourcesmalllanguage}. We see that in this case, AdaEDL and Max-Confidence-SPD increase the token rate by $43\%$ as compared to autoregressive decoding, while Base-SPD negatively impacts performance, reducing the token rate by $16\%$. 
    
    \subsection{Performance of AdaEDL across sampling temperatures}
\begin{figure}
        \centering
        \includegraphics[width=\linewidth]{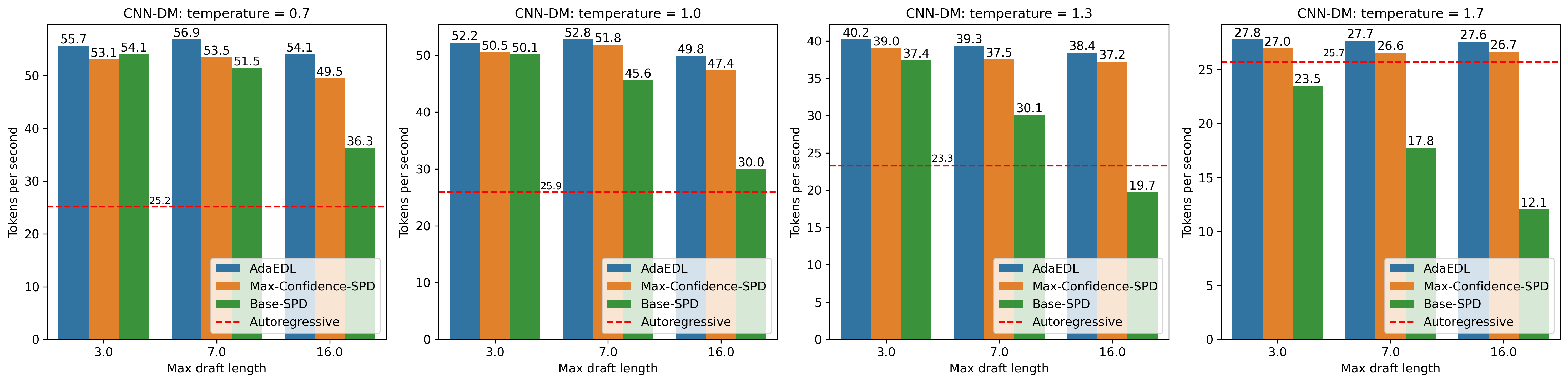}
        \caption{AdaEDL boosts token rate even in high temperature scenarios where Base-SPD is insufficient. Target model = Llama2-7B, draft model = Llama2-Drafter-115M, dataset = CNN-DM.}
        \label{fig:cnn-sampling-temp-bar-chart}
    \end{figure}

        In cases where the target distribution is difficult to predict, such as when the chosen sampling temperature is high, we see that Base-SPD is only able to produce modest gains in token rate - sometimes even resulting in poorer performance than autoregressive decoding. In Figure \ref{fig:cnn-sampling-temp-bar-chart} we see that as we increase the sampling temperature from $0.7$ to $1.7$ on the CNN-DM dataset, the token rate of Base-SPD drops $57\%$ (for draft length $3$), and ends up being lower than standard autoregressive decoding. On the other hand, even at high sampling temperatures, AdaEDL provides an $8\%$ boost in token rate over the autoregressive baseline. AdaEDL also consistently outperforms the other 3 decoding methods across lower sampling temperatures and across maximum draft length settings. We perform similar experiments on the Dolly-15k and WMT-19 datasets in Appendix \ref{appendix-sampling-temperature} Figures \ref{fig:dolly-sampling-temp-bar-chart}, \ref{fig:wmt-sampling-temp-bar-chart}, and see similarly consistent performance from AdaEDL. 
    
    % To justify the claim that AdaEDL would perform better than Max-Confidence-SPD when the probability distribution of tokens is no longer 
    
    % 1. Generate a bar chart for each of the temperature scenarios 
    % 2. For each temperature 
    %     Show the performance in DL3, DL7, DL16 
    % Generate more results between 0.3-0.8? 
    \subsection{Controllability and sensitivity of AdaEDL to hyperparameters} \label{sec:hyperparameters}
        \paragraph{Entropy factor ($\gamma$) } In all our experiments, we set $\gamma = 0.2$ in the equation $1 - \sqrt{\gamma H_{DM}(x)}$. We see that this simple approximation described in Appendix \ref{appendix-entropy-derivation}, when coupled with the dynamic stopping threshold strategy described in Algorithm \ref{alg:dynamic-lambda}, produces strong results. Further hyperparameter search may be possible to improve this value for a particular dataset and we observe that values in the range $\gamma \in (0, 1)$ typically work best. It may also be possible to dynamically update $\gamma$ using the observed target model distribution as the system runs. We defer these investigations to future works. 
        
        \paragraph{Threshold update hyperparameters ($\beta_1$, $\beta_2$, $\epsilon$, $\alpha$) } In all our experiments, we set $\beta_1 = 0.5$, $\beta_2 = 0.9$, $\epsilon = 0.01 $, and $\alpha = 0.9$, following \cite{zhang2024draftverifylossless}, for the threshold update step for both AdaEDL and Max-Confidence-SPD.

        \begin{figure}
            \centering
            \includegraphics[width=\linewidth]{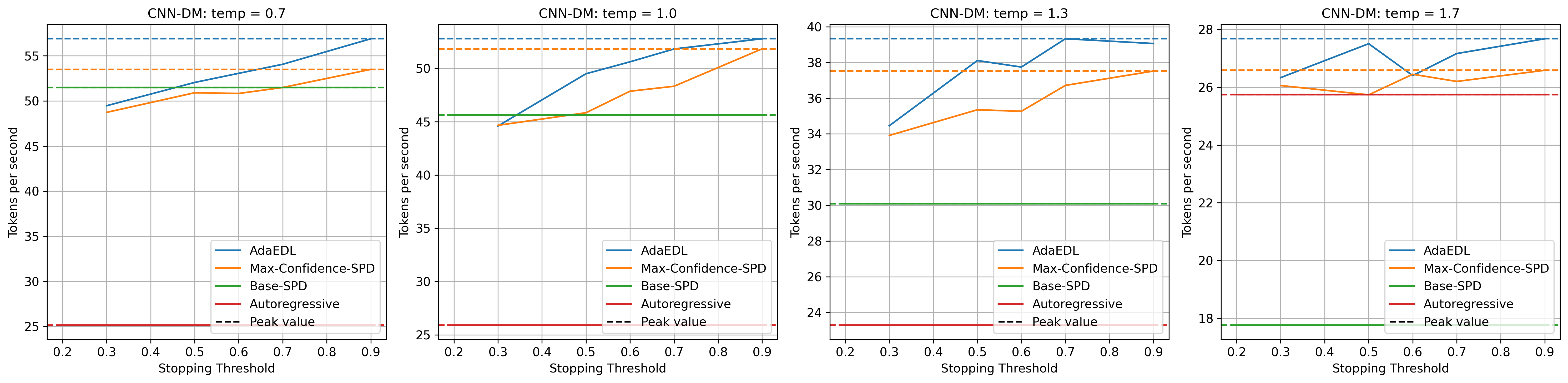}
            \caption{AdaEDL performs consistently better than Max-Confidence-SPD even at suboptimal $\lambda$. Target model = Llama2-7B, draft model = Llama2-Drafter-115M, dataset = CNN-DM, max draft length = $7$.}
            \label{fig:stopping-threshold-cnn-dl7}
        \end{figure}

        \paragraph{Stopping threshold ($\lambda$) } The most significant hyperparameter in adaptive draft length methods is the early draft-stopping threshold, $\lambda$. In all our experiments, $\lambda$ is updated dynamically according to the scheme described in Algorithm \ref{alg:dynamic-lambda}. Regardless, we find that both AdaEDL and Max-Confidence-SPD are sensitive to the initial choice of $\lambda$. We hypothesize that this may be due to the fact that many output generations are short in length, which may not result in enough drafting rounds for the system to converge to an optimal threshold. In Figure \ref{fig:stopping-threshold-cnn-dl7} we see that AdaEDL, even for sub-optimal $\lambda$ choices, consistently outperforms Max-Confidence-SPD when its $\lambda$ is also chosen sub-optimally. At the same time, AdaEDL performs better than this baseline if optimal $\lambda$ are chosen for both methods as we see marked by the dashed line and in Section \ref{sec:fast-finetuned-draft}. We conduct experiments across datasets (CNN-DM, Dolly-15k, WMT-19), sampling temperatures ($0.7$, $1.0$, $1.3$, $1.7$), and maximum draft length settings ($3$, $7$, $16$), and observe similar trends in Appendix \ref{appendix-stopping-threshold} Figures \ref{fig:stopping-threshold-cnn-dl7-new}, \ref{fig:stopping-threshold-dolly-dl7},  \ref{fig:stopping-threshold-wmt-dl7}, showing that AdaEDL can consistently boost token rate without the need for fine-grained hyperparameter search.

    % Show plots of varying the stopping threshold (temp = 1.0, WMT-19, Dolly-15k, CNN-DM) and show that blue curve is above orange curve even if they occasionally intersect 

% \subsection{Predictability}
%     \subsection{F1 score formulation}

%     \subsection{F1 score correlation}
%     Show how AdaEDL correlates more strongly with F1 score (wrt stopping threshold) as compared to max conf 

\section{Conclusion} \label{sec:conclusion}
    In this work, we present AdaEDL, an early stopping criteria for drafting in speculative decoding systems which uses the entropy of the draft model to estimate a lower bound on the current token's acceptance rate. We show the efficacy of this new method across datasets, sampling temperatures, draft lengths, and choice of target and draft models, whether fine-tuned or off-the-shelf. AdaEDL boosts the performance of existing speculative decoding systems while also enabling efficient usage of much larger draft models which, if finetuned in future works, could potentially result in even more impressive gains in token rate. AdaEDL is training and parameter-free, is not dependent on a given dataset, and also offers a relaxed choice in hyperparameters, making it a simple, plug-and-play improvement to a variety of pre-existing speculative decoding LLM systems.

\clearpage
% \section*{References}
% \printbibliography
\bibliographystyle{unsrtnat}
\bibliography{refs} % Replace with your .bib file

% References follow the acknowledgments in the camera-ready paper. Use unnumbered first-level heading for
% the references. Any choice of citation style is acceptable as long as you are
% consistent. It is permissible to reduce the font size to \verb+small+ (9 point)
% when listing the references.
% Note that the Reference section does not count towards the page limit.
% \medskip

% {
% \small

% [1] Alexander, J.A.\ \& Mozer, M.C.\ (1995) Template-based algorithms for
% connectionist rule extraction. In G.\ Tesauro, D.S.\ Touretzky and T.K.\ Leen
% (eds.), {\it Advances in Neural Information Processing Systems 7},
% pp.\ 609--616. Cambridge, MA: MIT Press.

% [2] Bower, J.M.\ \& Beeman, D.\ (1995) {\it The Book of GENESIS: Exploring
%   Realistic Neural Models with the GEneral NEural SImulation System.}  New York:
% TELOS/Springer--Verlag.

% [3] Hasselmo, M.E., Schnell, E.\ \& Barkai, E.\ (1995) Dynamics of learning and
% recall at excitatory recurrent synapses and cholinergic modulation in rat
% hippocampal region CA3. {\it Journal of Neuroscience} {\bf 15}(7):5249-5262.
% }

%%%%%%%%%%%%%%%%%%%%%%%%%%%%%%%%%%%%%%%%%%%%%%%%%%%%%%%%%%%%

\newpage % Ensure the appendix starts on a new page
\appendix

% \section{Appendix / supplemental material} \label{appendix-0}

% Optionally include supplemental material (complete proofs, additional experiments and plots) in appendix.
% All such materials \textbf{SHOULD be included in the main submission.}

\section{Acceptance rate variance for standard speculative decoding systems} \label{appendix-acceptance-rates}
\begin{figure} [htbp]
    \centering
    \begin{subfigure}[b]{1.0\textwidth}
        \includegraphics[width=\textwidth]{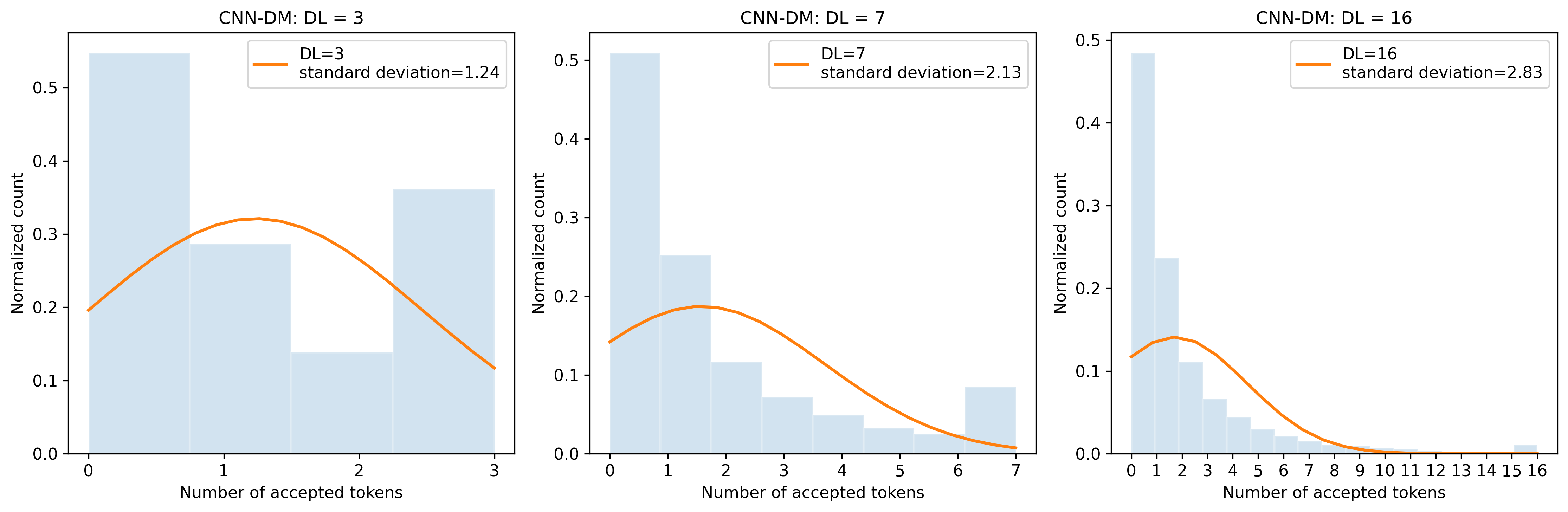}
        \caption{Dataset: CNN-DM}
        \label{fig:cnn-acceptance}
    \end{subfigure}
    \begin{subfigure}[b]{1.0\textwidth}
        \includegraphics[width=\textwidth]{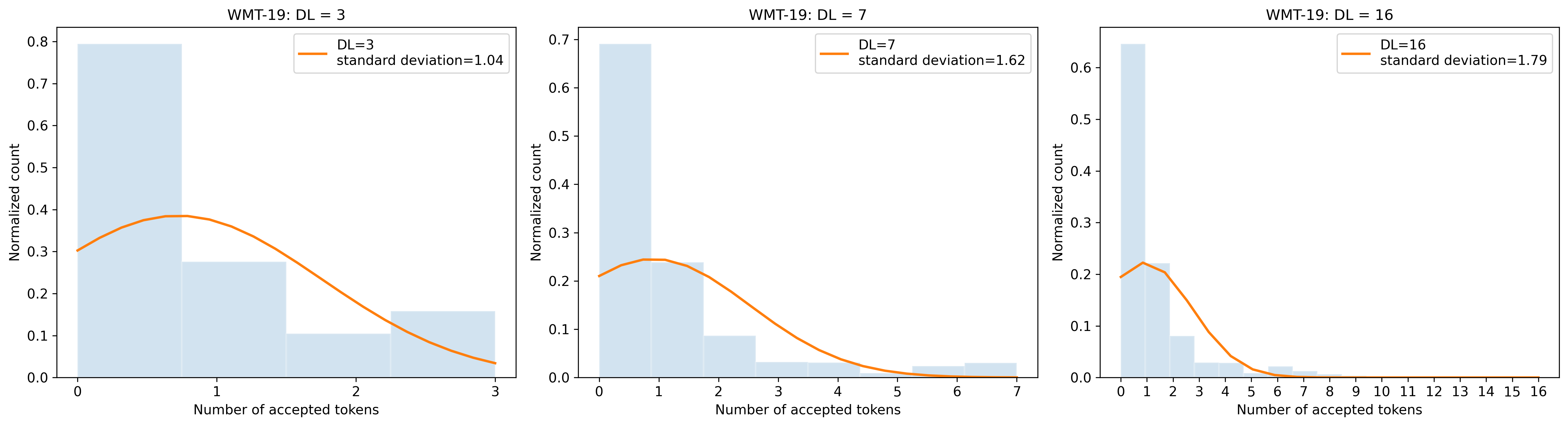}
        \caption{Dataset: WMT-19}
        \label{fig:wmt-acceptance}
    \end{subfigure}
    \begin{subfigure}[b]{1.0\textwidth}
        \includegraphics[width=\textwidth]{images/acc-rates-dolly.png}
        \caption{Dataset: Dolly-15k}
        \label{fig:dolly-acceptance-sub}
    \end{subfigure}
    \caption{Across datasets and draft lengths, the number of accepted tokens across drafting rounds displays a high variance, leading to under or over-utilization of the draft model in static draft length methods. Target model = Llama2-7B, draft model = Llama2-Drafter-115M, sampling temperature = $1.0$. }
\end{figure}

We see that across tasks like summarization (CNN-DM) in Figure \ref{fig:cnn-acceptance}, translation (WMT-19) in Figure \ref{fig:wmt-acceptance}, and creative writing (Dolly-15k) in Figure \ref{fig:dolly-acceptance-sub}, standard speculative decoding systems display a large variance in the number of tokens accepted per drafting round. This is observed across draft lengths $3$, $7$, and $16$. This effect is particularly pronounced in the CNN-DM and Dolly-15k datasets, in which we observe standard deviations of $\sim 2$ tokens even within a maximum draft length of only $7$ (i.e., $29\%$ standard deviation), opening up room for significant optimization. Experiments are conducted with target model chosen as Llama2-7B and a 115M draft model finetuned via direct alignment with the target model distribution \cite{goel2024directalignmentdraftmodel}. 

\section{Derivation of entropy-based draft-stopping criteria} \label{appendix-entropy-derivation}

Let $p_{DM}(x)$ and $p_{TM}(x)$ be the currently observed probability distributions of the draft and target model given some prefix $x_{<t}$. Following \cite{spd-leviathan}, the acceptance probability, $\beta$ of a token drafted from $p_{DM}(x)$ defined via rejection sampling is 

\[
     \beta = \mathbb{E}_{x \sim p_{DM}(x)}  \begin{cases*}
                                                1 & if  $p_{DM}(x) \leq p_{TM}(x)$ \\
                                                \frac{p_{TM}(x)}{p_{DM}(x)}  & if $p_{DM}(x) > p_{TM}(x) $
                                            \end{cases*}
\]

On discrete domains, the total variation distance between these distributions is defined in the following manner \cite{Levin2008MarkovCA} : 

$$ TVD (p_{DM} \| p_{TM}) = \dfrac{1}{2}\sum_x \left\lvert p_{TM}(x) - p_{DM}(x) \right\rvert $$

We see in \cite{spd-leviathan} that $TVD(DM \| TM) $ is related to the acceptance probability, $\beta$ as 

$$ \beta = 1 - TVD (p_{DM} \| p_{TM}) $$
$$ \implies TVD (p_{DM} \| p_{TM}) = 1 - \beta $$

Moreover, by Pinsker's inequality \cite{Rioul2024}, we may relate the total variation distance to the Kullback-Liebler divergence, $KLD( p_{DM} \| p_{TM} )$ 

$$ TVD (p_{DM} \| p_{TM})  \leq \sqrt{\frac{1}{2}KLD (p_{DM} \| p_{TM})} $$
$$ \implies 1 - \beta \leq \sqrt{\frac{1}{2}KLD(p_{DM} \| p_{TM})}  $$

Giving us, 
$$ 1 - \sqrt{\frac{1}{2}KLD(p_{DM} \| p_{TM})}  \leq \beta $$

Further, $KLD (p_{DM} \| p_{TM})$ relates to the cross-entropy $CE(p_{DM}, p_{TM})$ via 

$$ KLD(p_{DM} \| p_{TM}) = CE(p_{DM}, p_{TM}) - H_{DM} $$
where $H_{DM}$ is the entropy of the draft model distribution.

Now, let us note, while drafting, we do not yet have access to the target model distribution $p_{TM}(x)$. We do know, however, that since $KLD \geq 0$, we have that

$$ CE(p_{DM}, p_{TM}) \geq H_{DM} $$

In this work, we choose a linear approximation of $CE(p_{DM}, p_{TM})$ via a positive factor $\gamma' \in (0, 1) $ 

$$ CE(p_{DM}, p_{TM}) = (1 + \gamma') H_{DM} $$

This is motivated by the observation that in LLM systems, most of the variation seen in the cross-entropy between the draft and target model occurs due to the high entropy of the draft model. LLMs suitable to be target models follow the stable entropy hypothesis \cite{arora2023stableentropyhypothesisentropyaware} with reasonable generations lying in a narrow entropy band. 

Substituting this approximation, we have that 
\begin{align*}
    1 - \sqrt{\frac{1}{2}KLD \left(p_{DM} \| p_{TM}\right)} &\leq  \beta \\ 
    1 - \sqrt{\frac{1}{2}KLD \left(p_{DM} \| p_{TM}\right)} &= 1 - \sqrt{\frac{1}{2}\left(CE(p_{DM}, p_{TM}) - H_{DM}\right)} \\ 
        &\approx 1 - \sqrt{\frac{1}{2}\left( \left(1+\gamma'\right)H_{DM} - H_{DM}\right)} \\ 
        &= 1 - \sqrt{ \gamma H_{DM} } \\ 
\end{align*} 
% That is, 
% \begin{align*}
%         1 - \sqrt{ \gamma H_{DM} } &\leq \beta \\ 
% \end{align*}

Thus, we see that the value $1 - \sqrt{ \gamma H_{DM} } $ acts as an approximate \textit{lower bound} on the acceptance probability $\beta$. By stopping the drafting of a new token if our lower-bound estimate falls below a threshold $\lambda$, we attempt to ensure that the acceptance probability of the potential new token will be \textit{greater} than this threshold. If the acceptance probability does not meet our threshold, we choose not to draft the next token. 

Thus, a draft-stopping criteria 

$$1 - \sqrt{ \gamma H_{DM}(x) } < \lambda $$ 

implies that if drafting continues because $ 1- \sqrt{\gamma H_{DM}(x)} \geq \lambda$, then we have an approximate lower bound on the acceptance probability of the drafted token via $\lambda$, i.e., 

$$ \beta \geq \lambda $$

\subsection{Computational cost} \label{par:computational-cost}
 The cost of computing entropy is $O(N)$ on a single thread where $N$ is the size of the vocabulary. That said, this operation is highly parallelizable since the $N$ operations are independent. Thus, the overhead of AdaEDL is at most $O(N)$, but may be significantly reduced if implemented efficiently.  

\subsection{Impact of hardware chosen} \label{par:hardware-impact}
 An additional consideration is that the speedup achievable by a speculative decoding system depends on the cost of running the draft model, which depends on its size and the nature of the hardware it runs on. For example, a larger draft model may have a higher acceptance rate, but is also more expensive to run. An ideal system is one that balances these factors, taking into account the expected acceptance rate and computational cost incurred on a particular hardware. Future work that studies draft model cost across various processors would be valuable to designing such a system. 

\clearpage 
\section{Effect of target model sampling temperature} \label{appendix-sampling-temperature}
\begin{figure} [htbp]
    \centering
    \begin{subfigure}[b]{1.0\textwidth}
        \includegraphics[width=1\linewidth]{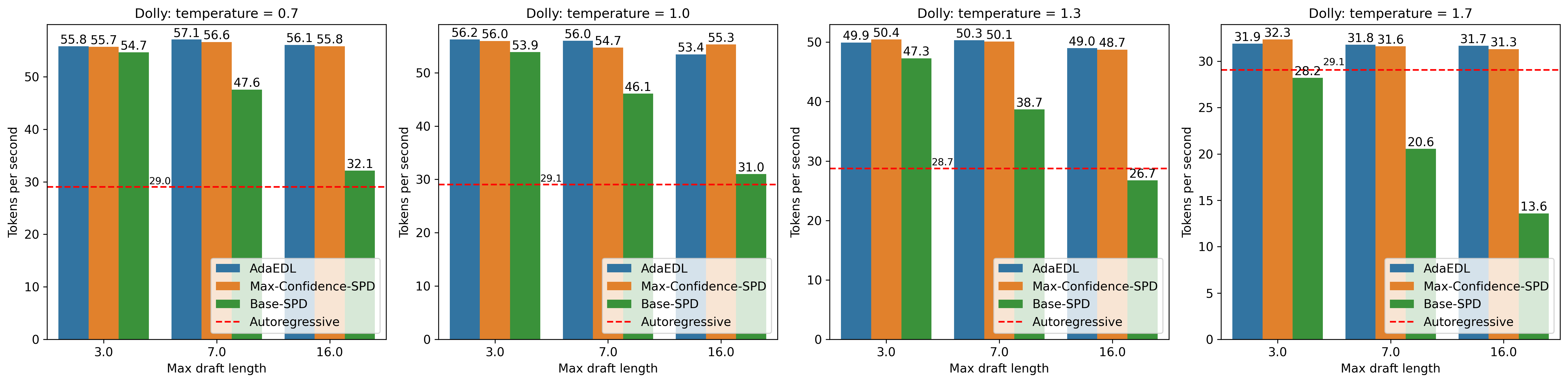}
        \caption{Dataset = Dolly-15k (creative writing)}
        \label{fig:dolly-sampling-temp-bar-chart}
    \end{subfigure}
    \begin{subfigure}[b]{1.0\textwidth}
        \includegraphics[width=1\linewidth]{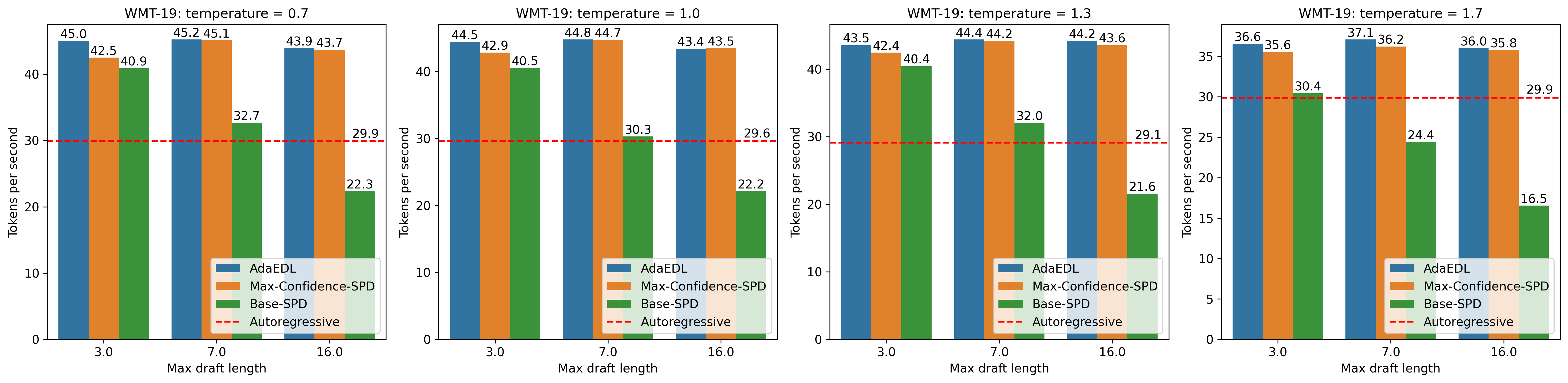}
        \caption{Dataset = WMT-19 (German-English translation)}
        \label{fig:wmt-sampling-temp-bar-chart}
    \end{subfigure}
    \begin{subfigure}[b]{1.0\textwidth}
        \includegraphics[width=\linewidth]{images/cnn-diff-temp-bar-chart.png}
        \caption{Dataset = CNN-DM (summarization)}
        \label{fig:cnn-sampling-temp-bar-chart-sub}
    \end{subfigure}
    \caption{AdaEDL boosts token rate even in high temperature scenarios where Base-SPD is insufficient. Target Model = Llama2-7B, Draft Model = Llama2-Drafter-115M.}
\end{figure}

AdaEDL consistently outperforms the other 3 decoding methods across \textbf{sampling temperatures}. This trend is reflected across maximum draft lengths $3$, $7$, $16$ and across datasets as seen for the Dolly-15k dataset (Figure \ref{fig:dolly-sampling-temp-bar-chart}), the WMT-19 dataset (Figure \ref{fig:wmt-sampling-temp-bar-chart}), and the CNN-DM dataset (Figure \ref{fig:cnn-sampling-temp-bar-chart-sub}).

\clearpage 
\section{Sensitivity to stopping thresholds} \label{appendix-stopping-threshold}
\begin{figure} [htbp]
    \centering
    \begin{subfigure}[b]{1.0\textwidth}
        \includegraphics[width=\linewidth]{images/cnn-diff-temp-dl-7.png}
        \caption{Dataset = CNN-DM (summarization)}
        \label{fig:stopping-threshold-cnn-dl7-new}
    \end{subfigure}
    \begin{subfigure}[b]{1.0\textwidth}
        \includegraphics[width=\linewidth]{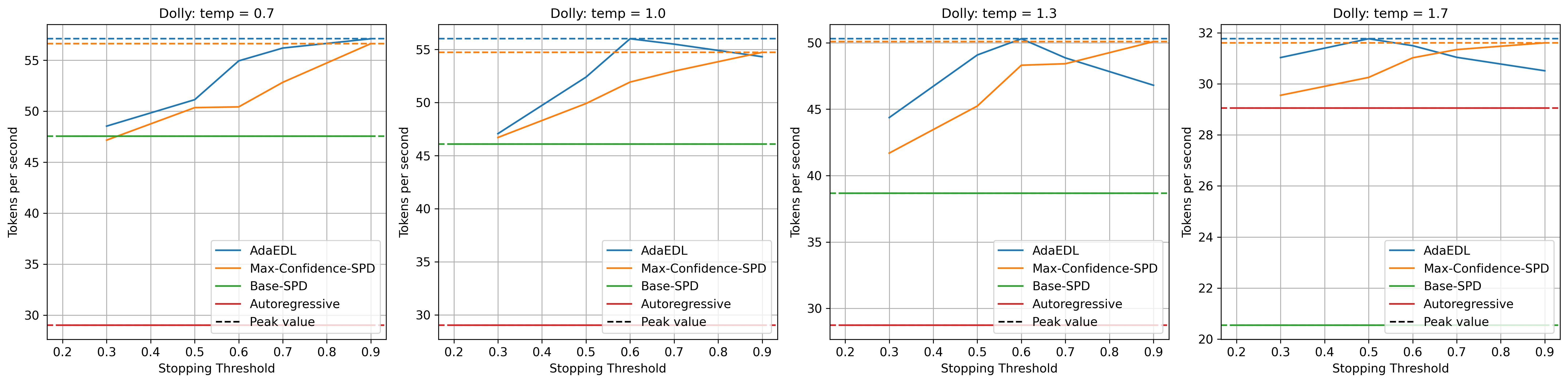}
        \caption{Dataset = Dolly-15k (creative writing)}
        \label{fig:stopping-threshold-dolly-dl7}
    \end{subfigure}
    \begin{subfigure}[b]{1.0\textwidth}
        \includegraphics[width=\linewidth]{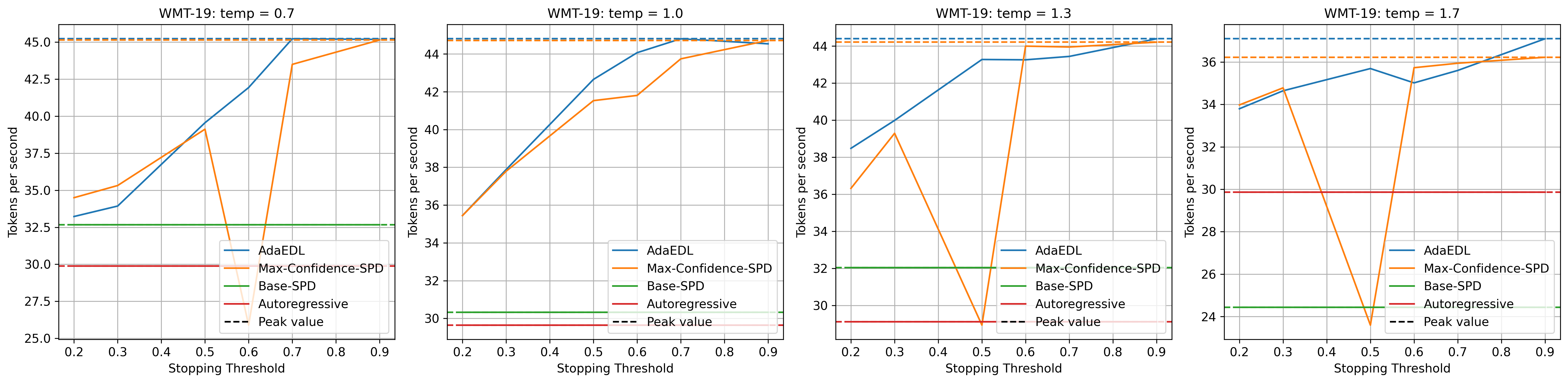}
        \caption{Dataset = WMT-19 (German-English translation)}
        \label{fig:stopping-threshold-wmt-dl7}
    \end{subfigure}
    \caption{AdaEDL maintains a margin over other methods even for sub-optimal stopping threshold choices. Max DL = $7$, TM = Llama2-7B, DM = Llama2-Drafter-115M. }
\end{figure}

We see that AdaEDL is less sensitive to the choice of the \textbf{stopping threshold} $\lambda$, outperforming Max-Confidence-SPD even when suboptimal $\lambda$ is chosen. This is reflected across temperatures and datasets as seen in Figures \ref{fig:stopping-threshold-cnn-dl7-new}, \ref{fig:stopping-threshold-dolly-dl7}, \ref{fig:stopping-threshold-wmt-dl7}.

\end{document}